\documentclass[11pt]{article}

\usepackage[T1]{fontenc}
\usepackage{graphicx}
\usepackage{amsmath, dsfont}
\usepackage{amssymb}
\usepackage{multirow}
\usepackage{booktabs}
\usepackage{tabularx}
\usepackage{cite}
\usepackage{algorithm}
\usepackage{algpseudocode}
\usepackage{subcaption}
\usepackage{soul}
\usepackage{lipsum}
\usepackage{doi}
\usepackage{float}

\begin{document}

\title{Objective-Induced Bias and Search Dynamics in Multiobjective Unsupervised Feature Selection}

\author{
Mathieu Cherpitel$^{1}$ \and Thomas Bäck$^{1}$ \and Martijn R. Tannemaat$^{2}$ \and Anna V. Kononova$^{1}$\\[0.8em]
$^{1}$LIACS, Leiden University, Leiden, The Netherlands\\
$^{2}$LUMC, Leiden University, Leiden, The Netherlands\\[0.6em]
}
\date{}

\maketitle

\begin{abstract}
Unsupervised feature selection is commonly formulated as a multiobjective optimisation problem that jointly optimises subset quality and subset size. Yet the behaviour of this formulation depends critically on the choice of evaluation objective, the direction of subset-size regularisation, and the initialisation strategy. We study these factors in a controlled setting using a synthetic dataset with known informative, redundant, and irrelevant feature types. Six formulations are compared by combining three evaluation objectives: accuracy, silhouette score, and PCA reconstruction loss with subset-size minimisation or maximisation. The results show that formulation strongly affects both search dynamics and the quality of the resulting Pareto front. Silhouette-based formulations exhibit a strong bias toward trivial low-cardinality solutions and remain weak proxies for predictive performance. In contrast, the proposed PCA loss objective produces compact subsets with test accuracy comparable to subsets obtained by directly optimising supervised accuracy. Overall, the study shows that the design of objective(s) is central to effective multiobjective unsupervised feature selection.

\end{abstract}

\section{Introduction}
Feature selection (FS) aims to identify a compact subset of features that preserves the information content of a dataset. This process is particularly beneficial for datasets with many features, which may lead to complexity issues due to the curse of dimensionality or result in long extraction times if computed over raw data, e.g., in time series. By removing potentially irrelevant and redundant features, FS lowers computational requirements and usually enhances the ability of a model to generalise to unseen data. By restricting the model to more informative features, it is less likely to mistake random noise in the training data for meaningful patterns.
The FS problem is inherently combinatorial, with a search space of size $2^d$, where $d$ denotes the number of features. Exhaustive search is therefore infeasible, even for moderate number of features, motivating heuristic optimisation strategies such as sequential selection methods and evolutionary algorithms~\cite{backEvolutionaryAlgorithmsParameter2023}. While sequential approaches iteratively construct subsets based on greedy criteria, evolutionary algorithms explore a broader set of candidate subsets, allowing for the consideration of more unique feature combinations.

Candidate feature subsets are typically evaluated using either unsupervised criteria, such as clustering quality, or supervised performance measures when labels are available. However, both types of objectives are strongly influenced by subset cardinality, which can lead to degenerate solutions if not explicitly controlled~\cite{handlFeatureSubsetSelection2006}. A widely adopted strategy to mitigate this issue is to formulate feature selection as a multiobjective optimisation problem, jointly optimising an evaluation objective and a subset-size regulariser (i.e. maximising accuracy while minimising subset-size)~\cite{emmanouilidisMultiobjectiveEvolutionarySetting2000}.
Despite its widespread use, this formulation introduces several underexplored design choices that can substantially influence optimisation behaviour. In particular, the direction of the subset-size regulariser (minimisation versus maximisation) and the initial distribution of subset cardinalities in the population can affect the search and determine which regions of the Pareto front are reachable under a limited evaluation budget. Moreover, in real-world datasets where the true structure of features is unknown, Pareto fronts are typically interpreted only through objective values, making it difficult to understand how different objectives relate to underlying feature types or redundancy structures.

To address these limitations, we propose a controlled experimental framework based on a synthetic dataset with an explicit feature taxonomy. The dataset is designed to include informative features, linearly and non-linearly redundant features, and multiple forms of noise, enabling direct inspection of the composition of selected subsets. This allows us to move beyond purely objective-based analysis and study how different optimisation choices affect both Pareto optimality and the structural properties of solutions in terms of feature content. Within this framework, we investigate multiobjective feature selection using three evaluation objectives: two unsupervised objectives, namely silhouette score and PCA reconstruction loss that we introduce, and a supervised classification objective used as a baseline objective. These objectives exhibit different sensitivities to subset cardinality, making them suitable for analysing objective-induced bias in multiobjective feature selection.

Our study makes three main contributions. First, we analyse how subset-size regularisation and initial population sampling strategies shape the structure of the Pareto front under different evaluation objectives. We show how these design choices determine which regions of the search space are explored under limited computational budgets, and, for both accuracy and silhouette-based objectives, how this affects the attainable trade-offs. Second, using a synthetic dataset with a known feature taxonomy, we analyse the composition of approximations of Pareto-optimal solutions in terms of informative, redundant, and irrelevant controlled features. This enables us to characterise how different objectives select feature types. Finally, we introduce an unsupervised objective, PCA loss which to the best to our knowledge has not been applied in this context, and analyse its behaviour under subset-size regularisation, addressing its cardinality bias. We evaluate it in terms of approximated Pareto front structure, feature composition, and alignment with downstream feature based classification performance.

\section{Related Work}
\subsection{Unsupervised Feature Selection}
Unsupervised feature selection (UFS) aims to identify informative subsets of features without access to target labels. The independence of UFS methods from known labels makes them applicable when targets are unknown or unreliable. By not relying on the target for selecting features, it reduces the risks of overfitting over the training data and  information leakage. In some cases, unsupervised selection methods have shown to achieve performance comparable to supervised approaches \cite{angSupervisedUnsupervisedSemiSupervised2016, haarComparisonSupervisedUnsupervised2019a} while having a higher potential at generalising to unseen data.

UFS methods can be broadly categorised into four main strategies~\cite{dwivediTaxonomyUnsupervisedFeature2024}. \textit{Wrapper methods} evaluate candidate feature subsets using a downstream unsupervised objective, such as clustering quality \cite{handlFeatureSubsetSelection2006}. \textit{Filter methods} rank or score features based on intrinsic data properties, such as similarity preservation \cite{mitraUnsupervisedFeatureSelection2002}, spectral structure \cite{heLaplacianScoreFeature2005b}, or variance and redundancy measures \cite{ferreiraUnsupervisedApproachFeature2012}. \textit{Embedded methods} integrate feature selection directly into the learning process while \textit{hybrid approaches} combine multiple strategies, for example by using filter methods to initialise or guide a wrapper-based search.

A common wrapper approach for unsupervised feature selection involves the use of a \textit{clustering algorithm} to assess whether the subset of interest exposes clear clusters in the data. They generally rely on clustering quality metrics such as the Silhouette score \cite{rousseeuwRousseeuwPJSilhouettes1987} or the Dabies-Boulin  index \cite{daviesClusterSeparationMeasure1979}. These metrics respectively measure both intra-cluster cohesion and inter-cluster separation and the ratio of within cluster scatter to between cluster separation. 

Despite offering an intuitive way to assess whether a reduced feature set preserves meaningful data structure by exposing clearly separated groups, the natural form of these objectives does not account for the \textit{induced cardinality bias}. In feature selection settings, good clustering scores can often be achieved with very few features, as reduced dimensionality may artificially enhance cluster contrast or suppress noise. As a result, these measures are sensitive to feature set cardinality and can favour small, potentially uninformative subsets. Optimisation based on such criteria may lead to trivial solutions, unless such bias is explicitly controlled \cite{handlFeatureSubsetSelection2006}. Additionally, their use as a wrapper approach can lead to significant computational cost as a clustering algorithm such as $k$-means must be repeated during the optimisation process. Furthermore, finding the right hyper-parameter values for the clustering algorithm can be difficult and may have a large impact on the perceived quality of the subset. For many clustering techniques that require the specification of the number of clusters, it has been shown that a dynamic number of clusters is preferable~\cite{dyFeatureSelectionUnsupervised}.

This subset dimensionality bias in feature selection is not unique to unsupervised settings. Even supervised objectives such as classification accuracy often exhibit weak sensitivity to the inclusion of irrelevant and redundant features, meaning that optimising accuracy alone may favour large feature subsets. To address this bias, two main strategies have been proposed: (i) modifying the objective function to account for subset size, for example by normalising by the subset size \cite{dyFeatureSelectionUnsupervised, kimEvolutionaryModelSelection2002}, or (ii) explicitly considering feature subset cardinality as a separate objective and solving the resulting multiobjective optimisation problem \cite{xueParticleSwarmOptimization2013, kozodoiMultiobjectiveApproachProfitdriven2019, hancerParetoFrontFeature2018}. In the rest of the paper, we focus on the latter.

\subsection{Multiobjective Feature Selection}
The transformation a single-objective problem into a multiobjective one by means of a helper objective is known as \textit{multiobjectivisation} \cite{knowlesReducingLocalOptima2001}. In feature selection, it typically involves optimising an \textit{evaluation objective} $f_1$ (e.g., accuracy, clustering quality or reconstruction error) jointly with a subset-size \textit{regulariser objective} $f_2$, which either penalises or rewards the number of selected features depending on the chosen direction. The resulting multiobjective formulation aims to explicitly control the cardinality bias inherent in many evaluation objectives by exposing the trade-off between solution quality and subset size.

Beyond the bias control, multi-objectivisation has been shown to offer several potential advantages. It can reduce the number of local optima and reshape the fitness landscape, making it easier to efficiently explore the solution space \cite{knowlesReducingLocalOptima2001}. It also introduces regions of incomparability between solutions, which can promote population diversity and, thus, improve exploration. While it is possible to keep the selection strategy single objective by combining the evaluation objective and regulariser objective into a scalar objective function, this requires careful selection of weights and does not offer the advantages of multi-objectivisation before mentioned. By approximating a set of Pareto-optimal solutions, multiobjective strategies provide a more complete representation of the trade-offs between the objectives \cite{maMultiobjectivizationSingleObjectiveOptimization2023}. As a result, multiobjective feature selection has become a widely studied and promising direction for feature selection, with numerous evolutionary approaches proposed in the literature \cite{jiaoSurveyEvolutionaryMultiobjective2024}.

\section{Problem Formulation}
We consider feature selection as a multiobjective optimisation problem over the space of all possible feature subsets. Let $X \in \mathds{R}^{n\times d}$ denote the original dataset containing $n$ samples and $d$ features. A candidate solution is then represented by a binary decision vector $x\in\{0, 1\}^d$ where each element $x_i$ acts as an indicator for the $i$-th feature. To evaluate the fitness of a candidate solution, we define the filtered dataset $X_x$ as the submatrix of $X$ consisting of the columns where $x_i = 1$. The \textit{multiobjective feature selection} (MOFS) problem is defined as the simultaneous minimisation of two competing objectives: 

\[
\min_{x\in\{0,1\}^d} (f_1(X_x), f_2(x))
\]

where $f_1(X_x)$ is an evaluation objective measuring the quality of the selected subset while $f_2(x)$ is a regularisation objective measuring the subset cardinality.

\section{Objectives}
\subsection{Silhouette Objective}
The silhouette objective evaluates the quality of clustering induced by the selected feature subset. Given a candidate solution $x$, a clustering algorithm is applied to the filtered dataset $X_x$. The silhouette score \cite{rousseeuwRousseeuwPJSilhouettes1987} measures the cohesion within clusters and the separation between clusters based on this subspace. For each sample $i$ in $X_x$, let $a_i(X_x)$ denote the mean distance between $i$ and all other samples in the same cluster and let $b_i(X_x)$ denote the minimum mean distance between $i$ and all samples in any other cluster. The silhouette coefficient for sample $i$ is defined as: 

\[
s_i(X_x) = \frac{b_i(X_x) - a_i(X_x)}{\max\{a_i(X_x), b_i(X_x)\}}
\]

The silhouette score of the filtered dataset $X_x$ is then given by the average over all samples: 

\[
s(X_x) = \frac{1}{n} \sum_{i=1}^{n} s_i(X_x)
\]

We use the implementation of the silhouette score provided by $sklearn$ \cite{pedregosaScikitlearnMachineLearning2018}. In this work, cluster labels are obtained using the $k$-means. For each subset $x$, the number of clusters $k$ is selected by maximising the silhouette score over a predefined range of candidate values. 
The silhouette score lies in $[-1,1]$, where higher values indicate better clustering quality. The negative of the silhouette score is then used as the evaluation objective, to be minimised.

\subsection{Accuracy Objective}
Although the primary focus of this work is unsupervised feature selection, we include classification accuracy as a reference objective to provide a supervised performance baseline. For a given subset $x$, a Random Forest (RF) classifier is trained on the filtered dataset $X_x$ and used to predict class labels. The classification accuracy is defined as the proportion of correctly classified samples:

\[
acc(X_x) = \frac{\text{number of correct predictions}}{\text{total number of predictions}}
\]

The accuracy takes values in $[0,1]$, where higher values indicate better predictive performance. The negative of accuracy is then used as the evaluation objective, to be minimised.

\subsection{PCA Loss Objective}
The Principal Component Analysis (PCA) loss objective evaluates the capacity of the filtered dataset $X_x$ to preserve the global variance and structural characteristics of the original dataset $X$. This objective is based on the hypothesis that a truly informative feature subset should act as a sufficient basis to linearly reconstruct the latent manifold of the full feature space.

First,  PCA is applied to the full dataset $X \in \mathbb{R}^{n \times d}$ to get the top $k$ principal components and form a target projection matrix $Z \in \mathbb{R}^{n \times k}$. This step is performed once as a pre-processing task, ensuring that the target $Z$ remains constant throughout the optimization process. Then, for a given candidate solution $x$, we train a multivariate linear regression model to predict the projection matrix from the filtered dataset $X_x$: $\hat{Z} = X_x W$, where $\hat{Z}$ is the predicted projection matrix from $X_x$ and $W$ the weights learned during training. The objective function then tries to minimise the Mean Square Error (loss) between $Z$ and $\hat{Z}$:

\[
f_{pca\_loss}(X_x) = \frac{1}{n}\sum_{i=1}^n||Z_i - \hat{Z_i}||^2
\]

While this approach may appear to be a complicated proxy for ranking features by PCA loadings, the reconstruction loss explicitly addresses feature redundancy. Traditional PCA loading analysis identifies features with the highest individual contribution to variance but treats each feature independently. Consequently, PCA often ranks highly correlated features at the top, leading to the selection of redundant subsets. In contrast, the PCA reconstruction loss evaluates the collective information of the subset. If a candidate feature is highly correlated with features already present in $x$, its contribution towards reducing the variance reconstruction loss will be negligible, as its variance is already ``explained'' by existing subset, even if its individual PCA loading is high.

The choice of PCA for defining the latent manifold $Z$ is motivated by the fact that principal components are linear combinations of the original features. Consequently, a multivariate linear regression model should be sufficient to recover this structure if the selected subset $X_x$ contains the necessary information. While non-linear dimensionality reduction techniques could help define $Z$, they would require a reconstruction model that accounts for non-linearity (e.g., neural networks), significantly increasing the computational overhead per fitness evaluation. Thus, within an evolutionary framework used for optimisation, the efficiency of linear reconstruction allows for a more careful exploration of the search space.

For this objective, a lower value indicates a better reconstruction of the latent space from the subset with a minimal value of 0. The evaluation objective is then defined as $f_{pca\_loss}(X_x)$ directly, to be minimised.

\subsection{Subset Size Regularisaser Objective}
The regulariser objective $f_2$ controls the cardinality of the selected feature subset, defined as:

\[
|x| = \sum_{i=1}^d x_i
\]

To fit the minimisation framework, we use $f_2(x) = |x|$ when minimising subset size and $f_2(x) = -|x|$ when maximising subset size.

\section{Setup}

\subsection{Synthetic Dataset Generation}
\label{sec:synth_data}
To better understand the behaviour of feature selection methods under different objective formulations, we adopt a controlled experimental setup based on a synthetic dataset with a known feature taxonomy. A central challenge in feature selection is the absence of a ground truth for the optimal feature subset. While supervised accuracy on a held-out test set is often used as an indirect evaluation criterion, it is dependent on the available labels and does not uniquely define an optimal subset. To address this limitation, we construct a three class classification problem composed of synthetic features with controlled properties and known types. The detailed generation procedures are provided in the supplementary material and generates the following feature types:

\begin{figure}[!t]
    \centering
    \includegraphics[width=0.75\linewidth]{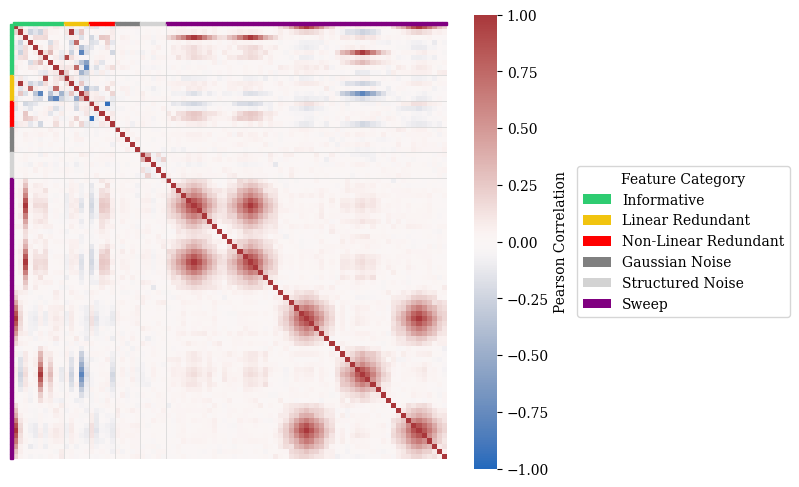}
    \caption{Correlation heatmap of the synthetic dataset used in this study.}
    \label{fig:synth_dataset_corr}
\end{figure}

\begin{itemize}
    \item \textbf{Informative}: sampled from clusters designed to separate the samples into classes, with added Gaussian noise to introduce variability. Features within each cluster are further linearly combined to induce covariance. They are generated using the \textit{make\_classification} function from scikit-learn \cite{pedregosaScikitlearnMachineLearning2018} and result in a set that is informative but does not allow perfect separation of the samples due to noise and random label flips.
    \item \textbf{Linear Redundant}: consist of linear combinations of informative features. Their purpose is to create redundancy and correlation in the dataset. This allows us to assess whether feature selection methods are capable of not selecting redundant features together.
    \item \textbf{Non-Linear Redundant}: introduce non-linear transformations of informative features. These features remain fully determined by the informative set but exhibit more complex relationships, testing the ability of methods to handle non-linear dependencies between features.
    \item \textbf{Gaussian Noise}: sampled independently from a standard normal distribution and are completely unrelated to the target. They serve as pure distractors, enabling the evaluation of a method’s robustness to irrelevant features.
    \item \textbf{Structured Noise}: exhibit some internal structure: samples can be separated but are generated independently of the target labels and with a different number of groups. Unlike Gaussian noise, they may appear informative due to their organization, allowing us to test whether methods are misled by structure that is not predictive of the task.
    \item \textbf{Sweep}: constructed by progressively perturbing individual informative features with varying levels of noise. This creates a collection of features with controlled correlation to the original signal, allowing us to evaluate how feature selection methods behave under gradual signal degradation and whether they can correctly prioritize stronger signals over weaker ones.
\end{itemize}
For features derived from informative variables, we record their lineage: the set of informative features used in their construction (for example, linear redundant feature 10 is a combination of informative features 1, 3, and 5), see Figure \ref{fig:synth_dataset_corr}. This enables analysis of selection behaviour, in particular explaining cases where the removal of an informative feature does not degrade supervised performance due to the presence of redundant features that preserves similar information content.

Although synthetic data generation approaches for feature selection have been proposed in the literature \cite{kimEvolutionaryModelSelection2002, kamalovSyntheticDataFeature2022}, existing methods do not typically provide non-linear combinations and lineage tracking. These limitations motivate the design of the dataset used in this study. The resulting correlation heatmap of the dataset used can be seen in Figure \ref{fig:synth_dataset_corr}.

While this generation procedure does not define a unique optimal feature subset (since certain redundant transformations may be more compact or predictive than the original informative features), it provides an intuitive reference solution consisting solely of the informative features, which we refer to as the \textit{naive ground truth}. Although the optimal choice among redundant features is not known, we assume that valid solutions should at least exclude purely noisy features, which, by construction, contain no information about the target.

\subsection{Initialisation Strategies}
In population-based evolutionary algorithms, the initialisation strategy refers to the method used to sample the initial population. 
Since these methods iteratively evolve a population, its initial location in the search space affects both convergence speed and the reachable regions. In this study, we investigate three different strategies:

\begin{itemize}
    \item \textbf{Binary Random Sampling:}
    A common approach for sampling the initial population for feature selection is independent binary random sampling or Bernoulli trials. Each feature is sampled from a uniform distribution and discretised using a threshold value $p$, corresponding to the probability that a feature is selected. In this setting, if the sampled number is lower than $p$, the feature is set to 1 (selected) and 0 (not selected) otherwise. With this initialisation strategy, the expected proportion of selected features in a solution is around $p$, and the resulting subset cardinalities are concentrated around this value across the initial population \cite{xuSegmentedInitializationOffspring2020}.
    When subset cardinality is included as an objective, the distribution of cardinalities in the initial population has a large impact on the following search dynamics. In particular, the interaction between this initial distribution and the direction of the subset-size regulariser (minimisation or maximisation) can bias exploration towards specific regions of the search space, potentially reducing coverage of other regions under a limited evaluation budget.
    For example, if the initial population is centred around a proportion of $p = 0.5$ selected features and the regulariser promotes larger subsets, regions corresponding to smaller subset sizes will be under-explored during the search.
    \item \textbf{Segmented Sampling}:
    To mitigate this limitation, \cite{xuSegmentedInitializationOffspring2020} proposed a segmented initialisation strategy designed to increase diversity along the cardinality axis. The population is divided into three equally sized sub-populations, each generated using a different selection probability (e.g., $p \in \{0.25, 0.5, 0.75\}$). The final initial population is obtained by concatenating these sub-populations. This results in a broader and more diverse distribution of subset cardinalities compared to standard binary random sampling.
    \item \textbf{Fixed Cardinality Sampling: }
    An alternative strategy used in the feature selection literature is fixed-cardinality sampling \cite{handlFeatureSubsetSelection2006}. In this approach, all individuals in the initial population are constrained to have the same number of selected features. While the subset size is fixed, diversity is introduced through variation in which features are selected.
    This removes variability along the cardinality axis in the initial population and focuses exploration entirely on feature composition. As a result, movement across different subset sizes depends entirely on the evolutionary operators and selection pressure induced by the optimisation process.
    This can introduce a limitation when the optimal regions of the search space lie far from the initial cardinality level. For example, if the population is initialised with very small subsets (e.g., one selected feature) but high-quality solutions require large subsets, reaching these regions may require many generations, depending on the strength of the selection pressure and the ability of variation operators to increase cardinality.
\end{itemize}

In this work, we compare how these initialisation strategies influence search dynamics, Pareto front structure, and the regions of the search space explored \textit{under different problem formulations}. This analysis aims to provide guidance on selecting an appropriate initialisation strategy based on the interaction between the evaluation objective and the subset-size regulariser.

\begin{table}[!t]
\centering
\caption{Overview of optimisation settings used in the experiments.}
\label{tab:eval_protocol}
\begin{tabular}{ll}
\toprule
\textbf{Component} & \textbf{Configuration} \\
\midrule
Optimiser & SMS-EMOA, implementation from~\cite{blankPymooMultiobjectiveOptimization2020}\\
Population size & 50 \\
Number of generations & 50 \\
Decision space & $x \in \{0,1\}^d$ \\
Evaluation Objectives ($f_1$) & Silhouette, Accuracy, PCA loss \\
Regulariser Objectives ($f_2$) & minimise size, maximise size \\
Initialisation & binary random, segmented, fixed cardinality \\
Test evaluation & Random Forest accuracy on held-out test set \\
\bottomrule
\end{tabular}
\end{table}

\subsection{Experimental Setup}
We evaluate multiobjective unsupervised feature selection across different combinations of evaluation objectives, subset-size regularisers, and initialisation strategies, while keeping optimisation parameters fixed. Each configuration corresponds to a unique combination of $f_1$, $f_2$ and initial population sampling strategy.
For each configuration, an independent optimisation process is executed, and both intermediate and final populations are recorded. The optimisation parameters shared across all configurations are listed in Table~\ref{tab:eval_protocol}. All configurations are evaluated on the same synthetic dataset described in Section \ref{sec:synth_data}. The optimisation process is performed exclusively on the training set, and all candidate solutions are evaluated using training data only. The final Pareto-optimal solutions are then assessed on the held-out test set using classification accuracy. For visual comparability across figures, subset size is reported in plots as the proportion of selected features: $|x|/d$, although the optimisation objective itself is defined in terms of the absolute cardinality $|x|$.

\section{Results}
\begin{figure}[!t]
    \centering
    \includegraphics[width=1\linewidth]{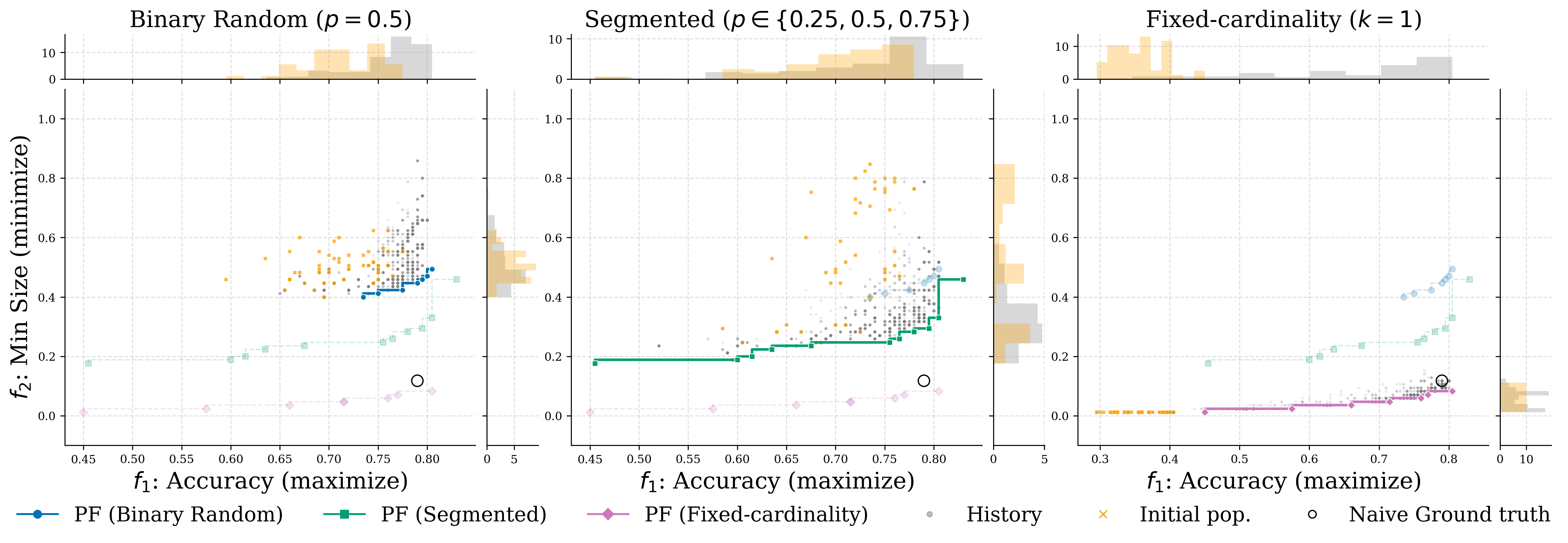}
    \caption{Search history, initial population location and found Pareto Front for simultaneous \textit{minimisation} of subset size and \textit{maximisation} of classification \textbf{accuracy} under three different initial population sampling strategies. Results from other strategies are shown with pale symbols for comparison.}
    \label{fig:front_accuracy_minsize}
\end{figure}

\begin{figure}[!t]
    \centering
    \includegraphics[width=1\linewidth]{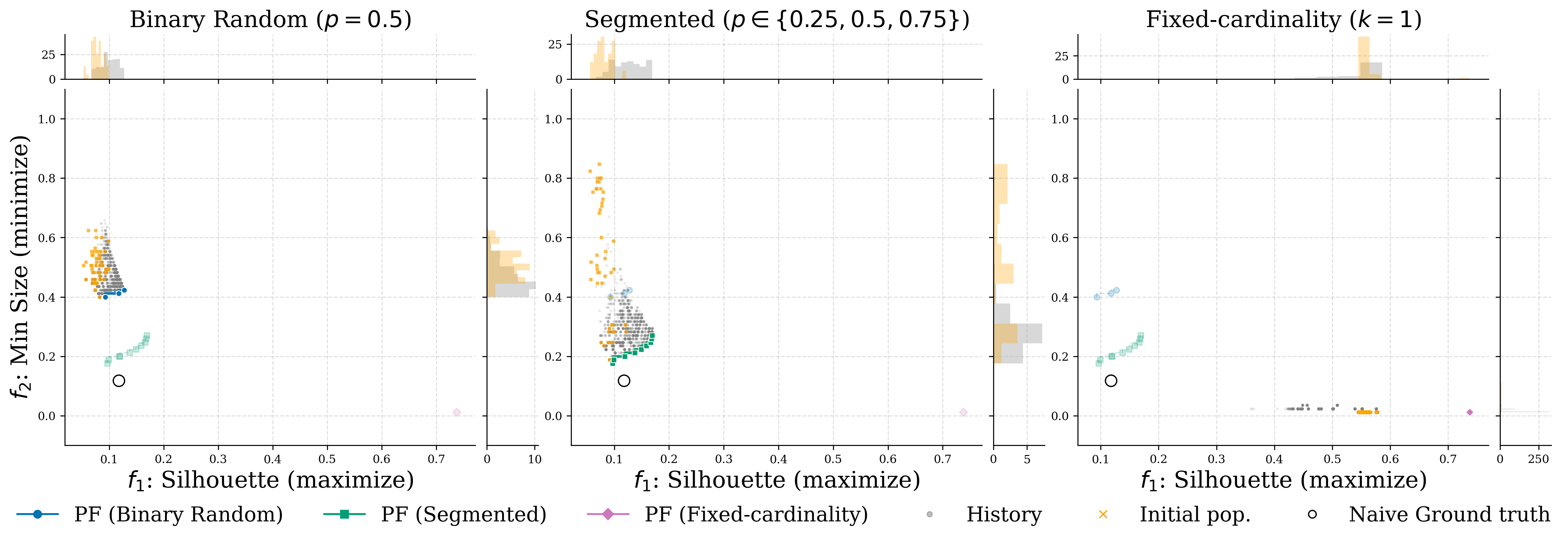}
    \caption{Search history, initial population location and found Pareto Front for simultaneous \textit{minimisation} of subset size and \textit{maximisation} of \textbf{silhouette score} under three different initial population sampling strategies. Results from other strategies are shown with pale symbols for comparison. Note the different range of the $f_1$ objective in comparison with Figure~\ref{fig:front_accuracy_minsize}.}
    \label{fig:front_silhouette_minsize}
\end{figure}
\subsection{Effects of Regularisation and Initialisation on the Search}
The standard approach in supervised MOFS involves the simultaneous maximization of classification accuracy and the minimization of subset cardinality. Since the starting point for feature selection is the hypothesis that there exists a more compact subset of features that preserves the information content of the dataset, it is intuitive to attempt to reduce the subset size in order to find optimal trade-offs between size and accuracy. As illustrated in Figure \ref{fig:front_accuracy_minsize}, this formulation allows for well-distributed Pareto Fronts and shows that the initialisation strategy has a strong effect on the explored search space and quality of the trade-offs found.

Intuitively, since the subset-size minimisation regulariser pushes the population toward smaller subsets, one might expect that sampling the initial population with relatively large or centrally distributed subset sizes would encourage exploration across the cardinality objective in search of a compact information-preserving subset. However, our results suggest otherwise. Given the high computational cost of evaluating objectives like Random Forest accuracy, the optimiser may fail to traverse the vast combinatorial space between high and low cardinalities within the allocated budget. Conversely, when the initial population is sampled with low cardinality (e.g., k=1), the multiobjective formulation ensures that, even in the presence of a ``minimize size'' regulariser, selection pressure from the accuracy objective drives the search toward larger subsets when they offer true gains in performance.

In a scenario where only a small portion of features are truly informative and non-redundant, our experiment showed better results with sampling the initial population with just one active feature.
Naturally, this effect is amplified by the fact that our synthetic dataset contains only a small proportion of truly informative features. As a result, initializing the population with single-feature subsets places the initial population closer to informative set compared to the other sampling strategies.

Although intuitive and, as demonstrated, beneficial for maximising accuracy, minimising the subset size is \textit{not} always the ideal formulation when searching for the best feature subset under every evaluation objectives. As demonstrated in Figure \ref{fig:front_silhouette_minsize}, pairing silhouette maximization with a size minimization regulariser can lead the search towards \textit{trivial solution}s of uninformative subsets. The silhouette score measures intra-cluster cohesion and inter-cluster separation, but it is inherently biased toward lower subset sizes~\cite{mierswaInformationPreservingMultiobjective2006}. In high-dimensional spaces, the ``curse of dimensionality'' compresses distance metrics, whereas very small subsets can artificially inflate the silhouette score by isolating features that form tight, distinct clusters by chance. This dimensionality bias has been previously described~\cite{mierswaInformationPreservingMultiobjective2006}, noting that clustering-based objectives often favour small, uninformative subsets unless the cardinality bias is explicitly controlled. When both the evaluation and regulariser objectives favour low cardinality, the search collapses toward the origin of the search space. As seen in the ``Fixed-cardinality ($k=1$)'' panel of Figure \ref{fig:front_silhouette_minsize}, the optimizer rapidly converges to trivial solutions. Despite achieving high silhouette scores, these subsets are far from the naive ground truth and fail to capture the informative structure of the synthetic taxonomy. Thus, silhouette score objective and clustering quality metrics in general \textit{should} be paired with a subset-size maximisation although unintuitive.

\begin{figure}[!t]
    \centering
    \begin{subfigure}{\textwidth}
        \centering
        \includegraphics[width=\linewidth]{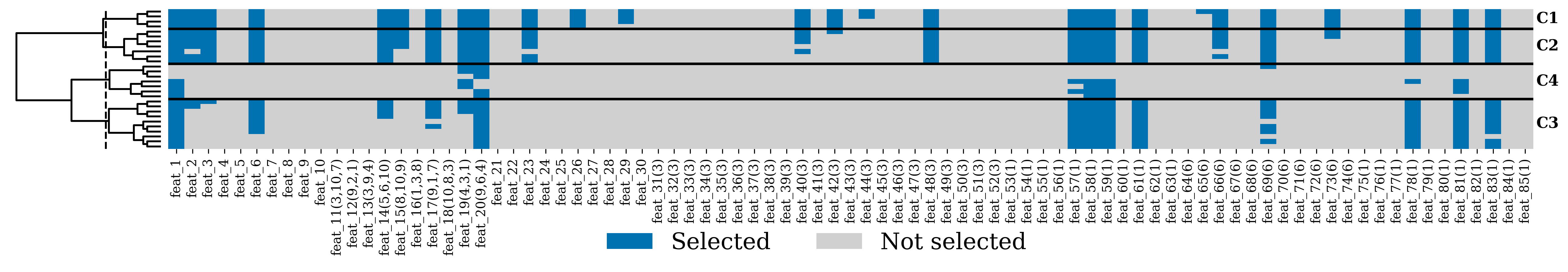}
        \caption{Feature composition across the Pareto front, clustered into groups C1 to C4.}
        \label{fig:composition_silhouette_maxsize}
    \end{subfigure}
    \begin{subfigure}{0.35\textwidth}
        \centering
        \includegraphics[width=\linewidth]{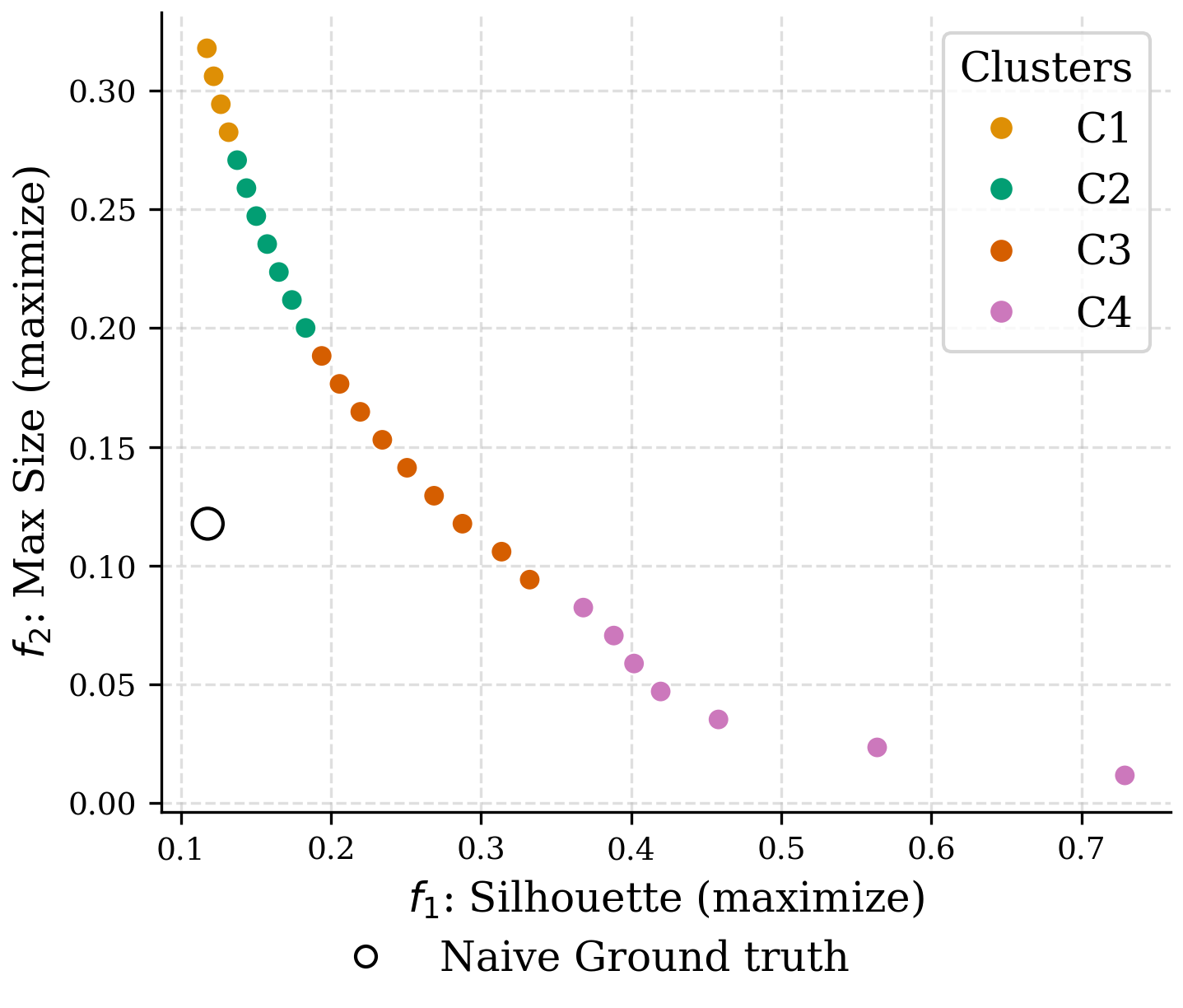}
        \caption{Clustered Pareto Front.}
        \label{fig:frontclust_silhouette_maxsize}
    \end{subfigure}
    \begin{subfigure}{0.35\textwidth}
        \centering
        \includegraphics[width=\linewidth]{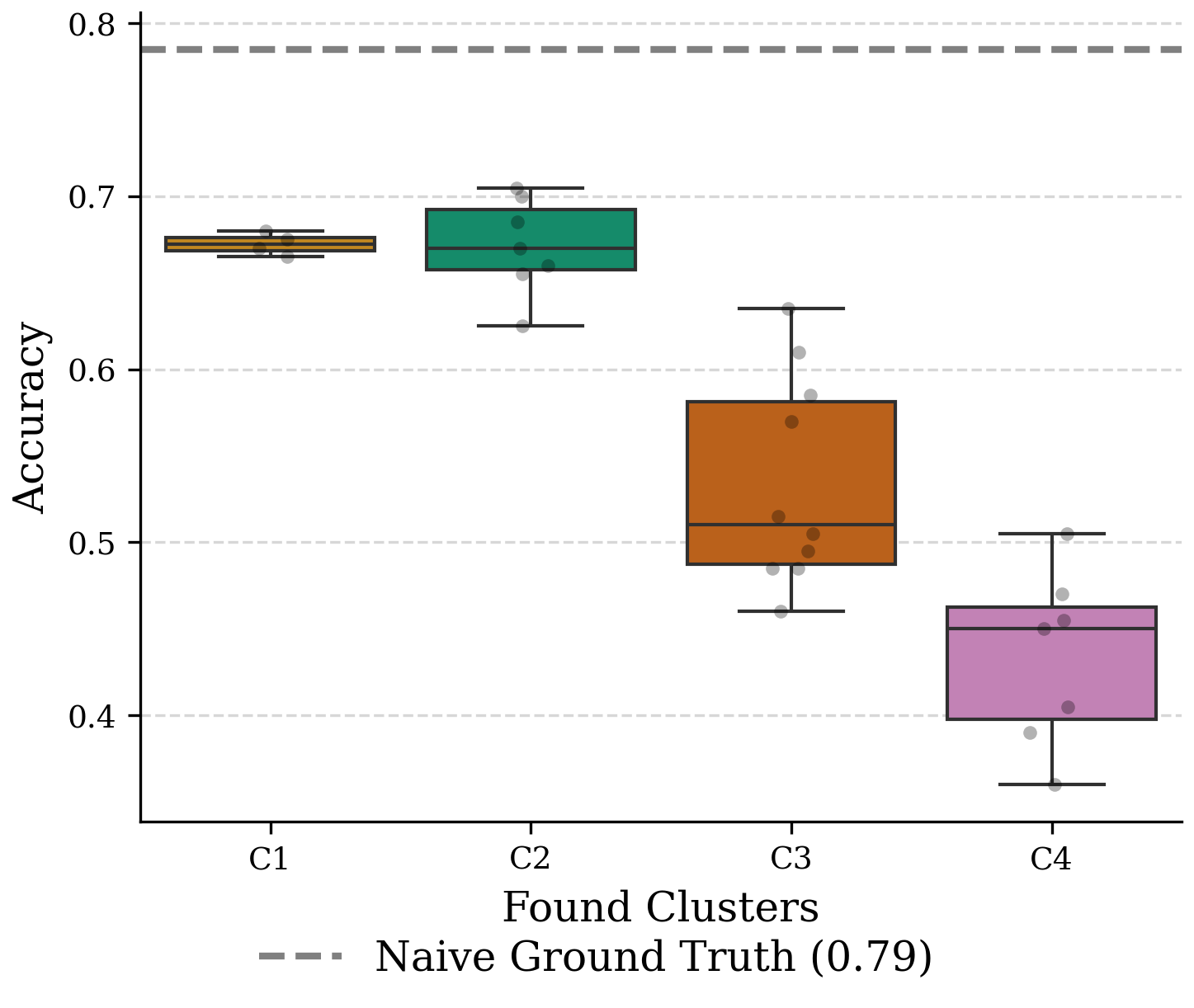}
        \caption{Accuracy distribution.}
        \label{fig:boxclust_silhouette_maxsize}
    \end{subfigure}
    \caption{Analysis of Pareto-optimal solutions for \textit{maximising} subset size while \textit{maximising} \textbf{silhouette score} with the fixed cardinality initialisation: (a) selected feature types per cluster; (b) mapping of clusters onto the found trade-offs; (c) distribution of classification performance for each group on a held-out test set.}
    \label{fig:combined_results_silhouette}
\end{figure}

\begin{figure}[!t]
    \centering
    \begin{subfigure}{\textwidth}
        \centering
        \includegraphics[width=\linewidth]{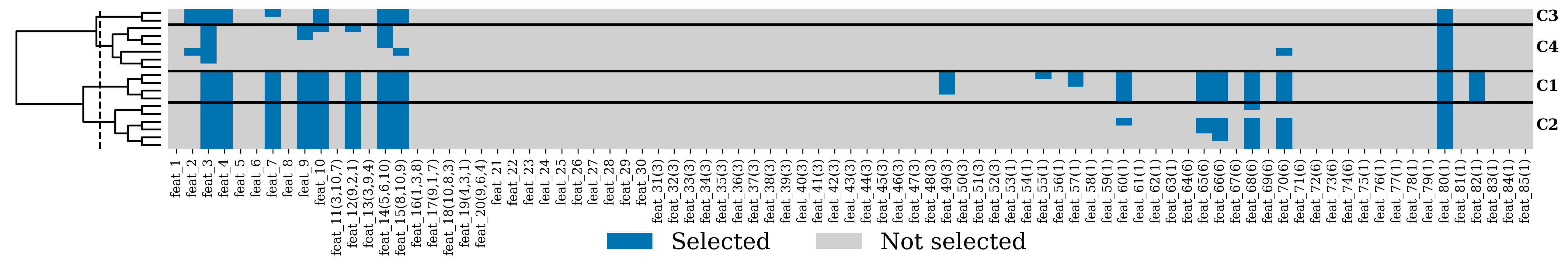}
        \caption{Feature composition across the Pareto front, clustered into groups C1 to C4.}
        \label{fig:composition_pca_minsize}
    \end{subfigure}
    \begin{subfigure}{0.35\textwidth}
        \centering
        \includegraphics[width=\linewidth]{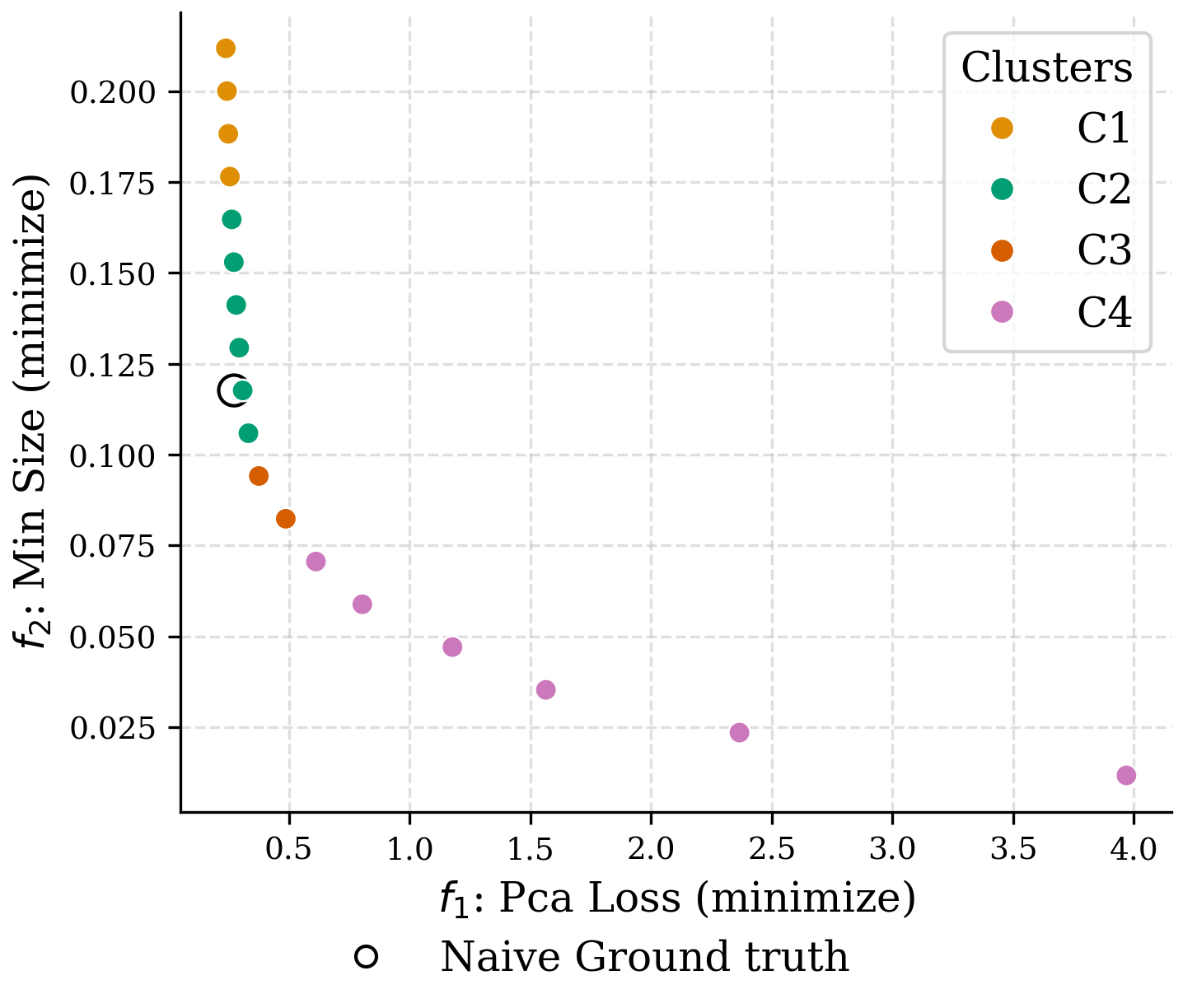}
        \caption{Clustered Pareto Front.}
        \label{fig:frontclust_pca_minsize}
    \end{subfigure}
    \begin{subfigure}{0.35\textwidth}
        \centering
        \includegraphics[width=\linewidth]{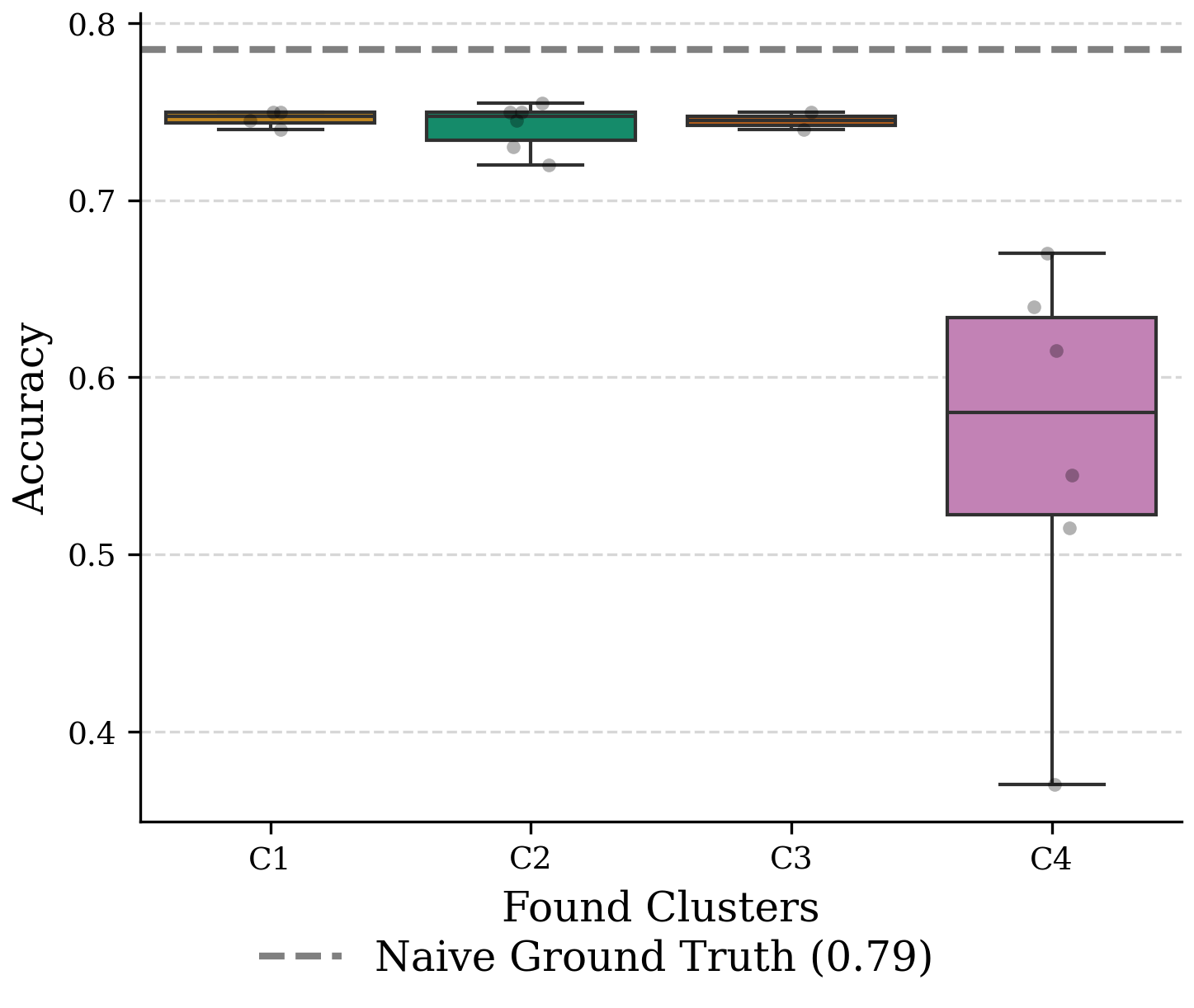}
        \caption{Accuracy distribution.}
        \label{fig:boxcluset_pca_minsize}
    \end{subfigure}
    \caption{Analysis of Pareto-optimal solutions for \textit{minimising} subset size while \textit{minimising} \textbf{PCA loss} with the fixed cardinality initialisation: (a) selected feature types per cluster; (b) mapping of clusters onto the found trade-offs; (c) distribution of classification performance for each group on a held-out test set.}
    \label{fig:combined_results_pca}
\end{figure}

\subsection{Composition of Pareto-optimal solutions}
We investigate the feature composition of solutions found on the found Pareto front for \textit{maximising} the silhouette score paired with the correct regularisation objective: subset-\textit{maximi\-sa\-tion}. Figure \ref{fig:composition_silhouette_maxsize} illustrates the selected features across the front. To compare different groups of solutions, we used a dendogram clustering (visible on the left) based on the selected features to split the solutions into groups C1 to C4. The obtained results show that this objective formulation, although correctly addressing the cardinality bias, leads to the selection of \textit{redundant} and even some \textit{irrelevant} features (e.g., 23, 26, 29). Feature 1 for example is selected multiple times, in almost all found solutions, inducing redundancy. Figure \ref{fig:frontclust_silhouette_maxsize} shows that despite the fact that groups C1 and C2 selected more features than what is present in the naive ground truth, some of the information was \textit{lost} in the selection, leading to lower classification accuracies seen in \ref{fig:frontclust_silhouette_maxsize}.
Furthermore, we observe in Figure \ref{fig:frontclust_silhouette_maxsize} that our naive ground truth is clearly \textit{dominated} by the found front but none of the dominating solutions achieve a higher classification accuracy (Figure \ref{fig:boxclust_silhouette_maxsize}), suggesting that this formulation of the silhouette score evaluation objective is \textit{not a good proxy for classification accuracy} on the considered synthetic dataset.

\subsection{PCA Loss as an Unsupervised Objective}
The solutions obtained on the Pareto front using the proposed PCA loss objective, combined with subset size \textit{minimisation}, are illustrated in Figure \ref{fig:composition_pca_minsize}. As in previous analyses, the solutions are grouped into four clusters according to the selected subsets. Inspection of clusters C1 and C2 indicates that a substantial proportion of the informative features has been successfully \textit{retained}. Although Feature 1, despite being informative, is not explicitly selected, its derived counterpart (Feature 80) is included, therefore its information content was \textit{preserved}. We notice that some redundancy persists. For instance, information associated with Feature 6 appears multiple times across certain solutions. As shown in Figure \ref{fig:frontclust_pca_minsize}, the Pareto front exhibits a clear trade-off between subset size and latent space reconstruction loss. Notably, the naive ground truth lies close to several identified solutions, indicating the method \textit{effectively approximates} near-optimal feature subsets. The performance distribution in Figure \ref{fig:boxcluset_pca_minsize} supports this: clusters C1, C2, and C3 closely approach the minimal ground truth, whereas cluster C4 features overly sparse subsets with lower performance. Overall, the proposed unsupervised objective shows \textit{strong potential} to preserve accuracy using compact subsets.

\begin{figure}[!t]
    \centering
    \includegraphics[width=1\linewidth]{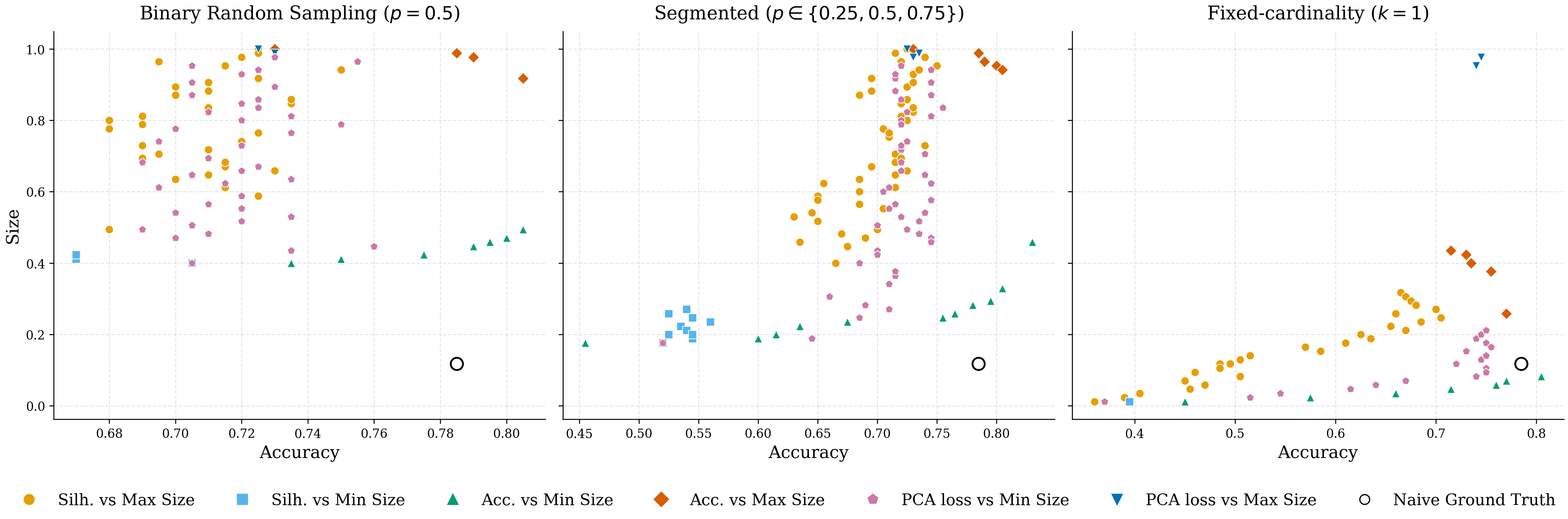}
    \caption{Pareto fronts of feature subsets under different MOFS formulations, evaluated on the test set (accuracy vs. subset size).}
    \label{fig:fronts_comparison}
\end{figure}

\begin{figure}
    \centering
    \includegraphics[width=0.97\linewidth]{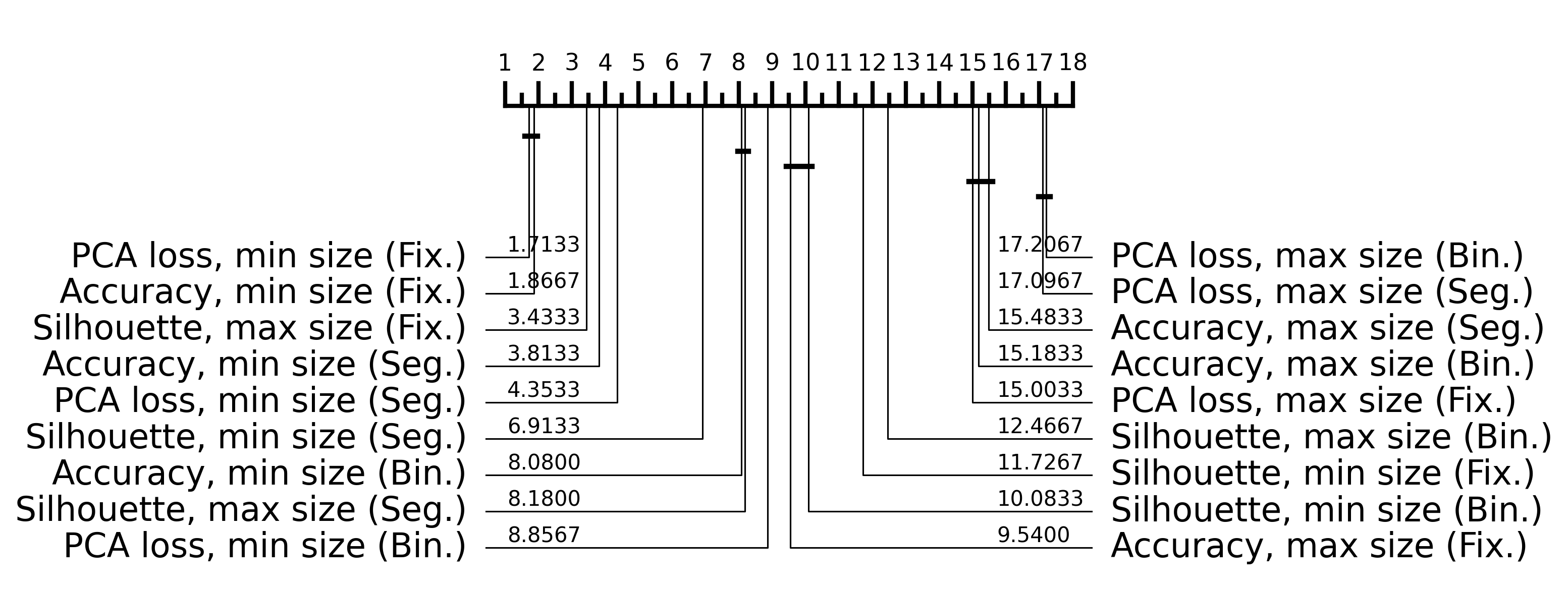}
    \caption{Critical difference plot ranking formulations by hypervolume indicator. Connected groups are not significantly different.}
    \label{fig:cd_fig}
\end{figure}

\subsection{Multiobjective feature selection problem formulations}
We compared six multiobjective formulations combining three evaluation objectives (accuracy, silhouette score, and PCA loss) with two regularization directions (subset-size minimization and maximization). The resulting Pareto fronts were evaluated on a held-out test set for the size–accuracy trade-off (Figure \ref{fig:fronts_comparison}). Directly optimizing for accuracy produced the strongest subsets. Unsupervised PCA loss yielded competitive solutions close to the supervised formulation, whereas silhouette-based approaches consistently underperformed. To \textit{validate} these results, we ran all formulations across 30 seeds on 5 synthetic datasets. Formulations were projected into accuracy–subset size space, evaluated via hypervolume indicators, and ranked using a post-hoc Friedman-Nemenyi test with Holm correction at $\alpha = 0.01$ (Figure \ref{fig:cd_fig}). The top-performing formulations were accuracy maximization and PCA loss, both using subset-size minimization with fixed cardinality sampling. While PCA loss does not outperform direct accuracy optimization, it is unsupervised, computationally cheaper, and significantly better than silhouette-based alternatives in our experiments. Evaluating these formulations on real-world datasets remains an essential next step. The remaining Figures for all formulations and a statistical analysis of feature categories in knee-point subsets which reveals the superiority of Accuracy and PCA loss formulations to discard noisy features are available in the supplementary Zenodo archive~\cite{zenodo_ppsn_supp}.

\section{Conclusion} 
In this work, we analysed multiobjective feature selection across different combinations of evaluation objectives, subset-size regularisation strategies, and initialisation schemes on a controlled synthetic dataset.
Our results show that problem formulation plays a decisive role in the quality of the selected subsets. In particular, the interaction between the evaluation objective and the regularisation direction strongly shapes the search behaviour and resulting Pareto front. Clustering-based objectives such as silhouette score can induce undesirable biases toward trivial solutions, whereas reconstruction-based objectives provide a more stable signal.
Notably, PCA loss emerges as a promising unsupervised alternative, achieving a favourable balance between subset compactness and predictive performance.
Future work should verify these findings on real-world datasets, explore a broader range of unsupervised objectives, and investigate whether more complex latent representations with non-linear reconstruction models can further improve performance.

\section*{Acknowledgements}
This work was supported by the ARISE-NMD project and funded by the Dutch Research Council (NWO) under the Open Technology Programme (project number 20852).

\bibliographystyle{splncs04}
\bibliography{references}

\end{document}